\documentclass{article}

    \PassOptionsToPackage{numbers, compress}{natbib}

    \usepackage[preprint]{neurips_2024}

\usepackage[utf8]{inputenc} 
\usepackage[T1]{fontenc}    
\usepackage{hyperref}       
\usepackage{url}            
\usepackage{booktabs}       
\usepackage{amsfonts}       
\usepackage{nicefrac}       
\usepackage{microtype}      
\usepackage{graphicx} 
\usepackage{xspace}

\usepackage{amsmath}
\usepackage{lipsum}
\usepackage{caption}
\usepackage{multirow}
\usepackage{makecell}
\usepackage{adjustbox}
\usepackage{wrapfig}
\usepackage{cleveref}
\usepackage{pifont}
\usepackage{marvosym}

\usepackage{multirow}
\usepackage[table,xcdraw]{xcolor}
\usepackage{makecell}
\usepackage{bm}

\definecolor{mygray}{gray}{0.95}

\definecolor{MyDarkGreen}{RGB}{45,155,45}

\title{VEnhancer: Generative Space-Time Enhancement \\ for Video Generation}

\author{Jingwen He$^{1,2}$ \quad Tianfan Xue$^{1}$ \quad Dongyang Liu$^{2}$ \quad Xinqi Lin$^{2}$ \quad Peng Gao$^{2}$ \\ 
\textbf{Dahua Lin}\textsuperscript{1} \quad \textbf{Yu Qiao}\textsuperscript{2}  \quad \textbf{Wanli Ouyang}\textsuperscript{1,2,\Letter}  \quad  \textbf{Ziwei Liu}\textsuperscript{3,\Letter}\\
\textsuperscript{1}The Chinese University of Hong Kong, \quad \textsuperscript{2}Shanghai AI Laboratory,\\
\textsuperscript{3}S-Lab, Nanyang Technological University \\
\url{https://vchitect.github.io/VEnhancer-project/}
}

\begin{document}

\maketitle
\footnotetext{\textsuperscript{\Letter} Corresponding authors.}
\begin{abstract}
We present \emph{VEnhancer}, a generative space-time enhancement framework that improves the existing text-to-video results by adding more details in spatial domain and synthetic detailed motion in temporal domain. Given a generated low-quality video, our approach can increase its spatial and temporal resolution simultaneously with arbitrary up-sampling space and time scales through a unified video diffusion model. Furthermore, VEnhancer effectively removes generated spatial artifacts and temporal flickering of generated videos.  
To achieve this, basing on a pretrained video diffusion model, we train a video ControlNet and inject it to the diffusion model as a condition on low frame-rate and low-resolution videos. To effectively train this video ControlNet, we design \textit{space-time data augmentation} as well as \textit{video-aware conditioning}.
Benefiting from the above designs, VEnhancer yields to be stable during training and shares an elegant end-to-end training manner.
Extensive experiments show that VEnhancer
surpasses existing state-of-the-art video super-resolution and space-time super-resolution methods in enhancing AI-generated videos. Moreover, with VEnhancer, exisiting open-source state-of-the-art text-to-video method, VideoCrafter-2, reaches the top one in video generation benchmark -- VBench.


\end{abstract}
\begin{figure}[t!]
  \centering
  \includegraphics[width = 0.95\linewidth]{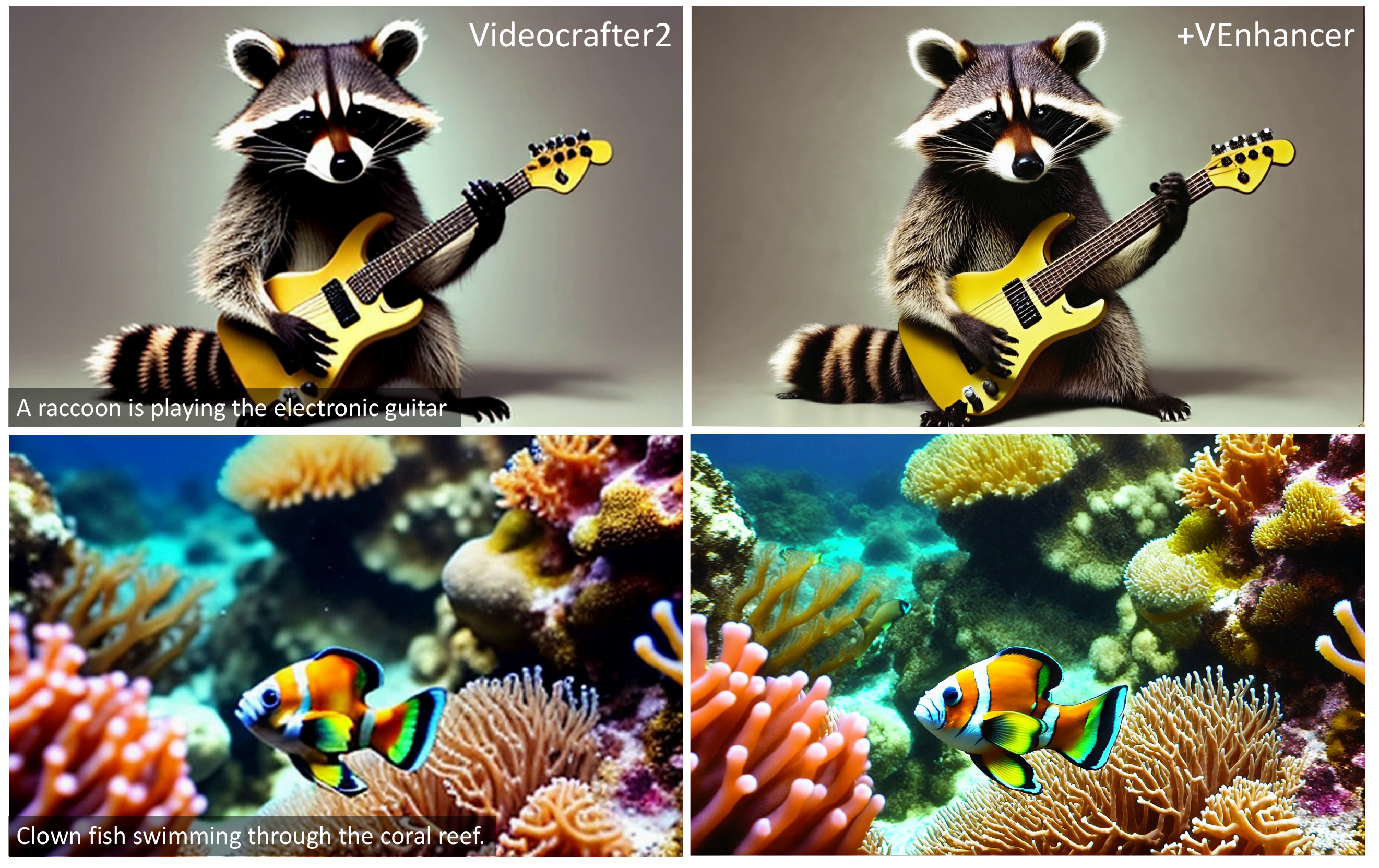}  
  \vspace{-8pt}
  \caption{
  The enhanced screenshots for AI-generated videos (from VideoCrafter-2 \cite{videocrafter2}). First row: VEnhancer is able to reconstruct the strings of guitar, and regenerate realistic fur of raccoon. Second row: VEnhancer could make the blurry background very sharp as well as enhance the color, which makes the whole picture vivid. \textbf{Zoom in for best view.}
  }
  \label{fig:teaser} \vspace{-2em}
\end{figure}

\section{Introduction}
With the advances of text-to-image (T2I) generation \cite{sd,sdxl,pixart,lumina} and large-scale text-video paired datasets \cite{webvid}, there has been a surge of progress in the field of text-to-video (T2V) generative models \cite{animatediff,gentron,imagenvideo,align,lavie,modelscope,videocrafter1,videocrafter2,gupta2023photorealistic,tuneavideo}. These developments enable users to generate compelling videos through textual descriptions of the desired content. 
One common solution~\cite{imagenvideo, align,lavie,svd,gupta2023photorealistic} to obtain high-quality videos is to adopt cascaded pipelines, which stacks several video diffusion models (DM), including T2V, temporal super-resolution (T-SR) and spatial super-resolution (S-SR) DMs. Such a pipeline significantly reduces computation cost when generating high-resolution and high-frame-rate videos. However, it poses several issues. First, it might be redundant and time-consuming to enhance videos in spatial and temporal axes \textit{separately} with different models, as spatial and temporal super-resolution are strongly correlated tasks. 
Second, the proposed diffusion-based T-SR/S-SR in \cite{align,imagenvideo,svd,lavie,diffbir} could only handle fixed interpolation ratio (\textit{i.e.}, predicting 3 frames between two consecutive frames) or fixed upscaling factors (\textit{i.e.}, $4\times$), thus the flexibility and functionality are limited. 
Third, training T-SR/S-SR using synthesized video pairs might result in inferior generalization ability as it could only generate low-level details without fundamentally understanding the semantics and structures of video contents. Therefore, such models only improve the spatial or temporal resolution with high fidelity, but struggle in modifying (\textit{i.e.}, generative enhancing) the original generated videos, such as eliminating video artifacts and flickering.

Another common approach mainly focuses on removing video artifacts and refining distorted content of generated videos.
One example is I2VGEN-XL~\cite{i2vgenxl}, which follows the idea of SDXL~\cite{sdxl} and uses a refinement model to remove visual artifacts and regenerate video content through a noising-denoising process~\cite{sdedit}.
I2VGEN-XL trains a diffusion model for refinement with large-scale high-quality video datasets captioned by short texts, leading to powerful regeneration ability.
Although this method has shown incredible performance in improving the stability of generated videos, it cannot increase the spatial and temporal resolution. More importantly, naive noising-denoising process will change the original video content significantly (\textit{i.e.}, sacrificing fidelity), which cannot always be acceptable in real-world applications.

To conclude, generative video enhancement methods have several limitations. First, cascaded temporal and spatial super-resolution diffusion models requires independent training, but their corresponding training datasets are basically the same (high-quality video datasets).
Thus, such design is both sub-optimal and inefficient during the inference.
Second, only fixed upscaling factors are supported for both spatial and temporal super-resolution, which limits their practicality.
Third, it struggles in obtaining a good balance between \textit{quality} and \textit{fidelity} for
generative enhancing (\textit{i.e.}, removing artifacts or flickering).

To this end, we propose \emph{VEnhancer}, a unified generative space-time enhancement framework which supports both spatial and temporal super-resolution with flexible space and time up-sampling scales. 
Additionally, it also remove visual artifacts and video flickering without severe drop in fidelity. To achieve this, we adopt a pretrained video diffusion model\footnote{https://modelscope.cn/models/iic/Video-to-Video/files} as the fixed generative video prior, supplying the generative ability for video enhancement.
To condition the generation on low frame-rate and low-resolution videos, we follow ControlNet \cite{controlnet} and copy the weights of multi-frame encoder and middle block from the generative video prior as the trainable condition network. 

Furthermore, to handle different up-sampling scales and reduce artifacts and flickering, we propose \textit{space-time data augmentation} to construct the training data.
In particular, at traininig stage, we randomly sample different downsampling factors (ranging from $1\times$~to $8\times$), different number skipped frames (ranging from $0$~to $7$), as well as different noise levels in noise augmentation~\cite{cascaded,imagenvideo}.
To ensure condition network be aware of the associated data augmentation applied to each input video, we also propose \textit{video-aware conditioning} that can realize different conditions across frames. To inject this condition, for key frame, the condition network takes additional input, including the multi-frame condition latents, the embeddings of the associated downscaling factor $s$ and noise level $\sigma$ by noise augmentation. 

Equipped with the above designs, VEnhancer yields stable and efficient end-to-end training. Extensive experiments have demonstrated VEnhancer's ability in enhancing generated videos (see visual results in Figure \ref{fig:teaser}, \ref{fig:vbench}). Besides,  
it outperforms state-of-the-art real-world and generative video super-resolution methods for spatial super-resolution only. Regarding space-time super-resolution for generated videos, VEnhancer also surpasses state-of-the-art methods as well as two-stage approach (frame interpolation + video super-resolution). 
Moreover, on the public video generation benchmark, VBench~\cite{vbench}, combined with VideoCrafter-2, our approach reaches the first place.


Our contributions can be summarized as below:
{
\vspace{-.6em}
\setlength{\itemsep}{0.0em}
\setlength{\parsep}{0.0em} 
\begin{enumerate}
\item We propose VEnhancer, an efficient approach for generative space-time super-resolution and refinement in a unified diffusion model for the first time. Inspired by ControlNet, we devise a \textit{video ControlNet} for multi-frame condition injection. Besides, effective data augmentation and conditioning schemes are proposed to assist the end-to-end model training.
\item VEnhancer is flexible to adapt to different upsampling factors for either spatial or temporal super-resolution, exceeding the limits of existing diffusion-based spatial or temporal super-resolution methods. Besides, it provides flexible control to modify the refinement strength for handling diversified video artifacts.
\item VEnhancer surpasses exisitng state-of-the-art video super-resolution methods and space-time super-resolution methods in enhancing generated videos.
With VEnhancer, existing text-to-video method VideoCrafter-2 \cite{videocrafter2} achieves the top one in VBench \cite{vbench} in both \textit{semantic} and \textit{quality}, outperforming professional video generation products, Gen-2 \footnote{https://runwayml.com/ai-tools/gen-2-text-to-video/} and Pika \footnote{https://pika.art/}.
\end{enumerate}
}

\section{Related Work}
\subsection{Video Generation}
Recently, there have been substantial efforts in training large-scale T2V \cite{videocomposer,imagenvideo,animatediff,gentron,gupta2023photorealistic,align,lavie,modelscope} models on large
scale datasets.
Some works \cite{align,lavie,modelscope} inflate a pre-trained text-to-image (T2I) model by inserting temporal layers and fine-tuning them or all parameters on video data, or adopts a joint image-video training strategy. 
In order to achieve high-quality video generation, \cite{imagenvideo,lavie,align} adopts multi-stage pipelines. In particular, cascaded video diffusion models are designed: One T2V base model that is followed by one or more frame interpolation and video super-resolution models. VideoLDM \cite{align}, LaVie \cite{lavie}, and Upscale-A-Video \cite{upscale} all develop the video super-resolution model based on $4\times$ sd (StableDiffusion \cite{sd}) upscaler, which has an additional downsampled image for conditioning the generation.
One drawback of this base model is losing quite a lot generative ability compared with T2I base models.
On the contrary, I2VGEN-XL follows SDXL \cite{sdxl} and uses noising-denoising process \cite{sdedit} to refine the generated artifacts. However, this strategy could improve stability but cannot increase the space-time resolution. VEnhancer is based on a generative video prior, and could address temporal/spatial super-resolution and refinement in a unified model.

\subsection{Video Enhancement}
\textbf{Video Super-Resolution.} Video Super-Resolution (VSR) is proposed to enhance video quality by upsampling low-resolution (LR) frames into high-resolution (HR) ones. Traditional VSR approaches\cite{vsr_transformer,basicvsr,basicvsr++,vsr_recurrent,vsr_tga,revisiting,vrt,recurrent,edvr,video_flow} often rely on fixed degradation models to synthesize training data pairs, which leads to a noticeable performance drop in real-world scenarios. To bridge this gap, recent advances\cite{investigating,mitigating} in VSR have embraced more diversified degradation models to better simulate real-world low-resolution videos. To achieve photo-realistic reconstruction, Upscale-A-Video\cite{upscale} integrates diffusion prior to produce detailed textures, upgrading VSR performance into next level. \textbf{Space-Time Super-Resolution.} Space Time Video Super-Resolution (STVSR) aims to simultaneously increase the resolutions of video frames in both spatial and temporal dimensions. Deep-learning based approaches\cite{starnet,fisr,zoom,videoinr} have achieved remarkable results on STVSR.  STARNet\cite{starnet} increases the spatial resolution and frame rate by leveraging the mutual information between space and time. FISR\cite{fisr} propose a joint framework with a multi-scale temporal loss to upscale the spatial-temporal resolution of videos. \cite{zoom} proposes a one-stage STVSR framework, which incorporates different sub-modules for LR frame features interpolation, temporal information aggregation and HR reconstruction. VideoINR\cite{videoinr} utilize the continuous video representation to achieve STVSR at arbitrary spatial resolution and frame rate. Although these methods obtain smooth and high-resolution output videos, but they fail in generating realistic texture details.

\section{Methodology}


\subsection{Preliminaries: Video Diffusion Models}
Our method is built on a video diffusion model ~\cite{i2vgenxl}, which is developed based on one of the latest text-to-image diffusion models, Stable Diffusion 2.1\footnote{https://huggingface.co/stabilityai/stable-diffusion-2-1}~\cite{sd}. Given an video $\textbf{x}\in\mathbb{R}^{F \times H \times W \times 3}$, the encoder $\mathcal{E}$ first encodes it into latent representation $\textbf{z}=\mathcal{E}(\textbf{x})$  frame-by-frame, where $\textbf{z}\in\mathbb{R}^{F \times H' \times W' \times C}$. Then, the forward diffusion and reverse denoising are conducted in the latent space. In the forward process, the noise is gradually added to the latent vector $\textbf{z}$ in total $T$ steps. And for each time-step $t$, the diffusion process is formulated as follows:
\begin{equation}
    \textbf{z}_t =\alpha_t \textbf{z} + \sigma_t \bm{\epsilon},
    \label{eq:diffusion}
\end{equation}
where $\bm{\epsilon}\in\mathcal{N}(\textbf{0},\textbf{I})$, 
and $\alpha_t$, $\sigma_t$ specify the noise schedule in which the corresponding log signal-to-noise-ratio ($\log[\alpha_t^2/\sigma_t^2]$) decreases monotonically with $t$. And at time-step $T$, $q(\textbf{z}_T) = \mathcal{N}(\textbf{0},\textbf{I})$.
As for backward pass, a diffusion model is used for iteratively denoising under the guidance of the text prompt $c_{text}$.
By adopting v-prediction parameterization \cite{vprediction}, the U-Net denoiser $f_{\theta}$ learns to make predictions of $\textbf{v}_t\equiv\alpha_t\bm{\epsilon}-\sigma_t \textbf{z}$. The optimization objective is simply formulated as:
\begin{equation}
\mathcal{L}_{LDM} = \mathbb{E}_{\textbf{z}, c_{text}, \bm{\epsilon} \sim \mathcal{N}(\textbf{0}, \textbf{I}), t }\Big[ \Vert \textbf{v} - f_\theta(\textbf{z}_{t},t, c_{text}) \Vert_{2}^{2}\Big].
\label{eq:revprocess}
\end{equation}
At the end, the generated videos are obtained through the VAE decoder: $\hat{\textbf{x}}=\mathcal{D}(\textbf{z})$.

\subsection{Architecture Design}
\begin{figure}[t!]
  \centering
  \includegraphics[width = 0.95\linewidth]{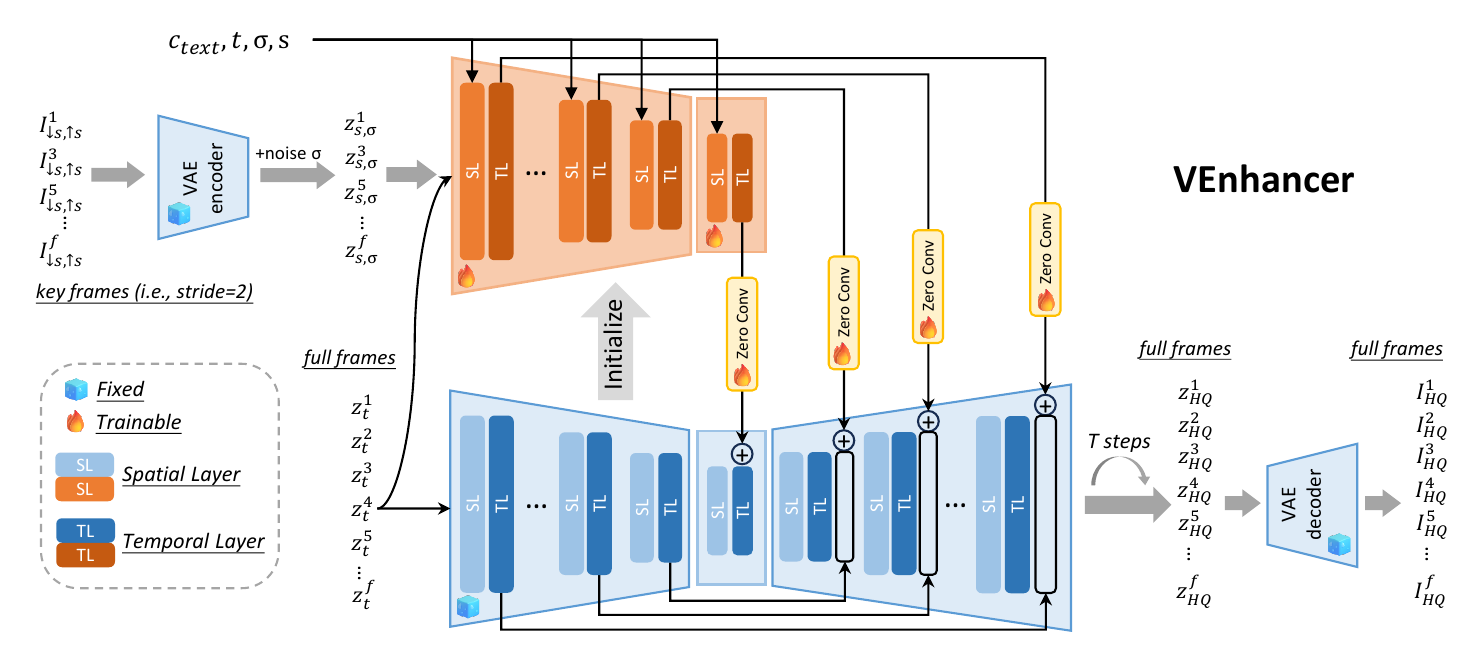}  
  \vspace{-0.5em}
  \caption{
  The architecture of VEnhancer. It follows ControlNet \cite{controlnet} and copies the architecures and weights of  multi-frame  encoder and middle block of a pretrained video diffusion model to build a trainable condition network. This \textit{``video ControlNet"} accepts low-resolution key frames as well as full frames of noisy latents as inputs. Also, the noise level $\sigma$ regarding noise augmentation and downscaling factor $s$ serve as additional network conditioning apart from timestep $t$ and prompt $c_{text}$. 
  }
  \label{fig:arch}
  \vspace{-1em}
\end{figure}

The architecture is built upon a video diffusion model.
This video diffusion model is able
to generate temporal-coherent content and high-quality texture details during iterative denoising. To upsample and refine a low-frame-rate and low-resolution videos in both spatial and temporal dimensions, the visual information should be incorporated into the video diffusion model carefully in order to obtain high-quality results without sacrificing fidelity significantly. High-quality generated videos stem from powerful generative models, while fidelity requires the algorithm to preserve the visual information in the input. Balancing the visual quality and fidelity is always challenging in generative model research. 
In this work, we follow ControlNet \cite{controlnet} to keep the pretrained video diffusion model untouched for preserving generative capability, but create a trainable copied network for effective condition injection. The architecture is illustrated in Figure \ref{fig:arch}. We will elaborate on our design carefully in subsequent paragraphs.

The pretrained video diffusion model follows the design of stacking a sequence of interleaved spatial and temporal layers within the 3D-UNet \cite{imagenvideo,align} architecture (blue blocks in Figure \ref{fig:arch}). Specifically, each spatial convolution layer (or attention layer) is followed by a temporal convolution layer (or attention layer). The spatial layers are the same as those in Stable Diffusion 2.1, including ResBlocks \cite{resnet}, self-attention \cite{attention} layers, and cross-attention layers.
The temporal convolution and attention layers are incorporated with their output layers initialized to zero and finetuned with video datasets. Specifically, the temporal convolution is one-dimensional convolution layer with a kernel size of 3, and the temporal attention is one-dimensional attention layer~\cite{modelscope}.
In this 3D-UNet, the video features that aligned by temporal layers in encoder will be skipped to the decoder, in which concatenation operation will be performed to combine skipped features with decoder features. 

To build the condition network, we make a copy (both the architectures and weights) of the multi-frame encoder and the middle block in 3D-UNet (orange blocks in Fig.~\ref{fig:arch}). This condition network also takes full frames of noisy latents as input, and outputs multi-scale temporal-coherent video features. These features will be injected into the original 3D-UNet through newly added zero convolutions (yellow blocks in Fig.~\ref{fig:arch}). The output features of the middle block in condition network will be added back to the features of the middle block in 3D-UNet. While for output features of encoder blocks in condition network, their features will be added to the skipped video features in 3D-UNet, which are also produced by encoder blocks. Such architecture design is consistent with the original ControlNet. The main difference is that video ControlNet has temporal convolution or attention layer positioned after spatial convolution or attention layer, and the multi-frame control signals are aligned by temporal layers.

\subsection{Space-Time Data Augmentation}
In this section, we discuss about how to achieve unified space-time super-resolution with arbitrary up-sampling space and time scales, as well as refinement with varying degrees. To this end, we propose a novel data augmentation strategy for both space and time axes. Details are discussed below.

\textit{\textbf{Time axis}}. Given a sequence of high-frame-rate and high-resolution video frames $I^{1:f} = [I^1, I^2, ..., I^f]$ with frame length $f$, we use a sliding window across time axis to select frames. The frame sliding window size $m$ is randomly sampled from a predefined set, ranging from $1$ to $8$. This corresponds to time scales from $1\times$ to $8\times$. Note that $1\times$ time scale requires no frame interpolation, thus the multi-task problem downgrades to video super-resolution. After the frame skipping, we obtain a sequence of key frames $I^{1:f:m} = [I^1, I^{1+m}, I^{1+2\times m}, ..., I^{f}]$.

\textit{\textbf{Space axis}}. Then, we perform spatial downsampling for these obtained key frames. Specifically, the downscaling factor $s$ is ramdomly sampled from $[1, 8]$, which represents $1\times\sim8\times$ space super-resolution. When $s=1$, there is no need to perform spatial super-resolution. 
All frames in one sequence are downsampled with the same downscaling factor $s$. Thus, we arrive at low-frame-rate and low-resolution video frames: $I_{\downarrow s}^{1:f:m}$. In practice, we should upsample them back to the original spatial sizes by bilinear interpolation before being passed to the networks, so we obtain $I_{\downarrow s,\uparrow s}^{1:f:m}$. 
Note that each space or time scale corresponds to different difficulty level, and thus the sampling is not uniform. Particularly, we set sampling probabilities of scales $4\times$ and $8\times$ based on a ratio of $1:2$, which is determined by their associated scale values.  

Then, we use the encoder part of a pretrained variational autoencoder (VAE) $\mathcal{E}$ to project the input sequence to the latent space frame-wisely:
\begin{equation}
z_{s}^{1:f:m} = [\mathcal{E}(I_{\downarrow s,\uparrow s}^{1}), \mathcal{E}(I_{\downarrow s,\uparrow s}^{1+m}), \mathcal{E}(I_{\downarrow s,\uparrow s}^{1+2\times m}), ..., \mathcal{E}(I_{\downarrow s,\uparrow s}^{f})].
\end{equation}

\textit{\textbf{Noise augmentation in latent space}}. At this stage, we conduct noise augmentation to noise the latent condition information in varying degrees in order to achieve controllable refinement. This noise augmentation process is the same as the diffusion process \eqref{eq:diffusion} used in the video diffusion model. Specifically, the condition latent sequence is corrupted by:
\begin{equation}
    z_{s, t'}^{1:f:m} = \alpha_{t'} z_{s}^{1:f:m} + \sigma_{t'} \epsilon^{1:f:m},
    \label{eq:noise_aug}
\end{equation}
where $\alpha_{t'}$, $\sigma_{t'}$ determine the signal-to-noise-ratio at time-step $t'$, and $t'\in \{1,...,T'\}$. Note that the pretrained video diffusion model adopts 1,000 steps ($T=1000$ in Eq. \eqref{eq:diffusion}). While the noise augmentation only needs to corrupt the low-level information, $T'$ is set to $300$ empirically.
For more intuitive denotation, we use $\sigma$ instead of $t'$. Finally, we arrive at $z_{s, \sigma}^{1:f:m}=\mathcal{E}(I_{\downarrow s,\uparrow s})_{\sigma}^{1:f:m}$.

The whole process of space-time data augmentation is summarized as follows:
\begin{equation}
    I^{1:f} \rightarrow I^{1:f:m} \rightarrow I_{\downarrow s}^{1:f:m} \rightarrow I_{\downarrow s,\uparrow s}^{1:f:m} \rightarrow \mathcal{E}(I_{\downarrow s,\uparrow s})^{1:f:m} \rightarrow \mathcal{E}(I_{\downarrow s,\uparrow s})_{\sigma}^{1:f:m}.
\end{equation}


\subsection{Video-Aware Conditioning}
\begin{wrapfigure}[21]{r}{0.45\textwidth}
\centering
\vspace{-18pt}
\includegraphics[width=0.45\textwidth]{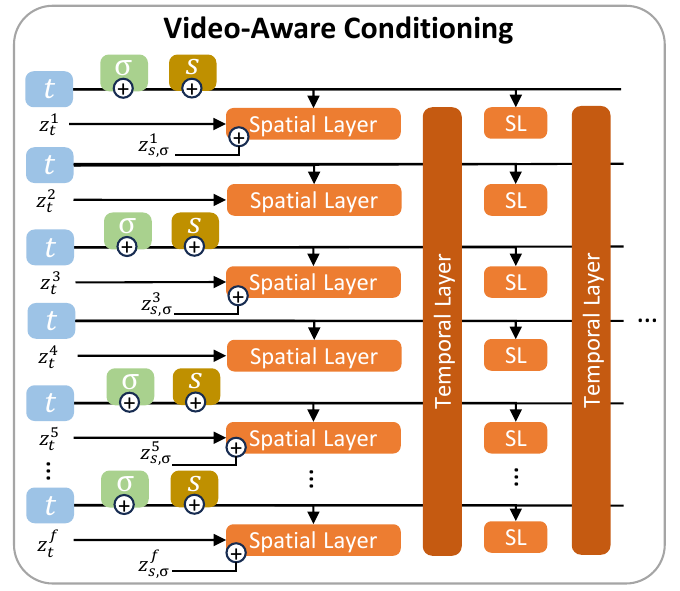}
  \caption{Video-aware conditioning. For frame that has condition image as input (key frame), we add it to the first layer of video ControlNet. Besides, the embeddings of noise level $\sigma$ and downscaling factor $s$ are added to the existing $t$ embedding, which will be broadcast to all spatial layers.}
  \label{fig:conditioning}
\end{wrapfigure}

Besides data augmentation, the corresponding conditioning mechanism should also be designed in order to boost the model training and avoid averaging performance for different space or time scales and noise augmentation. In practice, the condition latent sequence $z_{s, \sigma}^{1:f:m}$, the corresponding downscaling factor $s$, and augmented noises $\sigma$ are all considered as for conditioning. Please refer to Fig.~\ref{fig:conditioning} for more intuitive demonstration.

Given the synthesized condition latent sequence $z_{s, \sigma}^{1:f:m}$, we use one convolution layer with zero-initialization $\texttt{Conv}_{\texttt{zero}}$ for connecting it to video ControlNet. Specifically, we have:
\begin{align}
   & f_{out}^{1:f} = \texttt{Conv}(z_{t}^{1:f}),  \\ 
   & f_{out}^{1:f:m} = \texttt{Conv}(z_{t}^{1:f:m})  + \texttt{Conv}_{\texttt{zero}}(z_{s, \sigma}^{1:f:m}), 
\end{align}
where $\texttt{Conv}$ is the first convolution layer in video ControlNet, $z_{t}^{1:f}$ and $z_{t}^{1:f:m}$ denote the full frames and key frames of noisy latents at timestep $t$, respectively.
Note that $\texttt{Conv}$ and $\texttt{Conv}_{\texttt{zero}}$ share the same hyper-parameter configuration (\textit{i.e.,}kernel size, padding, et.al.), 
As it is shown, only key-frame features in video ControlNet will be added with the condition features, while others remain unchanged. This strategy enables progressive condition injection as the weights of $\texttt{Conv}_{\texttt{zero}}$ grows from zero starting point. 

For conditioning regarding downscaling factor $s$ and noise augmentation $\sigma$, we incorporate them to the existing time embedding in video ControlNet. Specifically, for timestep $t$, sinusoidal encoding \cite{ddpm,sd,attention} is used to provide the model with a
positional encoding for time. Then, one $\texttt{MLP}$ (two linear layers with a SiLU \cite{silu} activation layer in between) is applied. Specifically, we have: 
\begin{align}
    &t_{pos} = \texttt{Sinusoidal}(t), \\
    &t_{emb} = \texttt{Linear}_\texttt{t}^{(2)}(\texttt{SiLU}(\texttt{Linear}_\texttt{t}^{(1)}(t_{pos}))), \\
    &t_{emb}^{1:f} = \texttt{Repeat}(t_{emb}, f),
\end{align}
where $t_{emb}^{1:f}$ is obtained by $\texttt{Repeat}$  $t_{emb}$ by $f$ times in the frame axis. This time embedding sequence will be broadcast to all ResBlocks in video ControlNet for timestep injection. 

Also, we elucidate the conditioning for noise augmentation. As mentioned in Eq.\eqref{eq:noise_aug}, noise augmentation shares the same way as diffusion process, but with much smaller maximum timestep (\textit{i.e.,} $T'=300$). Therefore, we reuse the encoding and mapping for timestep $t$ in diffusion process. After this, we add a linear layer with zero initialization (denoted as $\texttt{Linear}_\texttt{zero}$).
To conclude, we have:
\begin{align}
    &\sigma_{pos} = \texttt{Sinusoidal}(\sigma), \\
    &\sigma_{emb} = \texttt{Linear}_{\texttt{zero},\sigma}(\texttt{Linear}_\texttt{t}^{(2)}(\texttt{SiLU}(\texttt{Linear}_\texttt{t}^{(1)}(\sigma_{pos})))). 
\end{align}
To achieve video-aware conditioning, we add $\sigma_{emb}$ only to the key frames. So $\sigma_{emb}$ is repeated $k$ times, where $k$ is the number of key frames. The video-aware controlling is presented as follows:
\begin{align}
    t_{emb}^{1:f:m} = t_{emb}^{1:f:m} + \sigma_{emb}^{1:k},
\end{align}
where the addition operation is performed frame-wisely for key frames.

Regarding downscaling factor $s$, the corresponding encoding, mapping and controlling are similar as above. 
In particular, we newly introduce one \texttt{MLP}, in which the output layer is zero-initialized. The video-aware conditioning is performed as:
\begin{align}
    &s_{pos} = \texttt{Sinusoidal}(s), \\
    &s_{emb} = \texttt{Linear}_{\texttt{zero},\texttt{s}}(\texttt{SiLU}(\texttt{Linear}_{\texttt{s}}(s_{pos}))), \\
    &s_{emb}^{1:k} = \texttt{Repeat}(s_{emb}, k), \\
    &t_{emb}^{1:f:m} = t_{emb}^{1:f:m} + s_{emb}^{1:k}.
\end{align}
With our proposed \textit{space-time data augmentation} and \textit{video-aware conditioning}, VEnhancer can be  well-trained in an end-to-end manner, and yields great performance for generative enhancement.

\section{Experiments}
\textbf{Datasets.}
We collect around 350k high-quality and high-resolution video clips from the Internet to constitute our training set. 
We train VEnhancer on resolution $720\times1280$ with center cropping, and the target FPS is fixed to $24$ by frame skipping.
Regarding test dataset, we collect comprehensive generated videos from state-of-the-art text-to-video methods \cite{modelscope,lavie,videocrafter2,show1}. Practically, we select videos with large motions and diverse contents. This test dataset is denoted as AIGC2023, which is used to evaluate VEnhancer and baselines for video super-resolution and space-time super-resolution tasks.
For evaluation on VBench, all generated videos based on the provided prompt suite are considered, resulting in more than 5k videos.

\textbf{Implementation Details.}
The batch size is set to 256. AdamW
\cite{adamw} is used as the optimizer, and the learning rate is set to $10^{-5}$.
During training, we dropout the text prompt with a probability of 10\%.
The training process lasts about four days with 16 NVIDIA A100 GPUs. During inference, we use 50 DDIM \cite{ddim} sampling steps and classifier-free guidance (\textit{cfg}) \cite{cfg}.

\textbf{Metrics.}
Regarding evaluation for video super-resolution and space-time super-resolution on AIGC2023 test dataset, we use both image quality assessment (IQA) and video quality assessment (VQA) metrics. As there is no ground-truth available, we can only use non-reference metrics. Specifically,  MUSIQ \cite{musiq} and DOVER \cite{dover} are adopted. Moreover, we refer to video generation benchmark,
VBench \cite{vbench}, for more comprehensive evaluation. 
Specifically, we choose \textbf{Subject Consistency} (\textit{i.e.,} whether the subject remains consistent), \textbf{Motion smoothness} (\textit{i.e.,} how smooth the video is), \textbf{Aesthetic Quality}, and \textbf{Imaging Quality} for evaluation. Regarding evaluation for video generation, we consider all 16 evaluation dimensions from VBench. 

\subsection{Comparison with Video Super-Resolution Methods}
For video super-resolution, 
VEnhancer is compared with the state-of-the-art real-world video super-resolution method, RealBasicVSR \cite{investigating}, and the state-of-the-art generative video super-resolution method, LaVie-SR \cite{lavie} (super-resolution). 

\begin{table}[!h]
    \centering
    \small
    \vspace{-1em}
    \caption{Quantitative comparison for video super-resolution ($4\times$) on AIGC2023 test dataset. {\color{red}\textbf{Red}} and {\color{blue}blue} indicate the best and second best performance. The top 3 results are marked as \colorbox{mygray}{gray}.}
    \label{tab:sr_x4}
\begin{tabular}{c|cc|cccc}
\toprule
             & DOVER↑                          & MUSIQ↑                           & \makecell{Imaging\\Quality}                & \makecell{Aesthetic\\Quality}              & \makecell{Subject\\Consistency}            & \makecell{Motion\\Smoothness}              \\ \hline
LaVie-SR \cite{lavie}       & {\color[HTML]{0B5FD1} 0.8427}  & {\color[HTML]{0B5FD1} 55.8428}  & {\color[HTML]{0B5FD1} 0.5481}  & {\color[HTML]{0B5FD1} 0.6692}  & {\color[HTML]{0B5FD1} 0.9562}  & \cellcolor[HTML]{F2F2F2}0.9710 \\
RealBasicVSR \cite{investigating} & \cellcolor[HTML]{F2F2F2}0.8252 & \cellcolor[HTML]{F2F2F2}50.5978 & \cellcolor[HTML]{F2F2F2}0.5401 & \cellcolor[HTML]{F2F2F2}0.6622 & \cellcolor[HTML]{F2F2F2}0.9555 & {\color[HTML]{0B5FD1} 0.9729}  \\
Ours         & {\color[HTML]{FF0000} \textbf{0.8498}}  & {\color[HTML]{FF0000} \textbf{56.6113}}  & {\color[HTML]{FF0000} \textbf{0.5872}}  & {\color[HTML]{FF0000} \textbf{0.6728}}  & {\color[HTML]{FF0000} \textbf{0.9624}}  & {\color[HTML]{FF0000} \textbf{0.9787}}  \\ \bottomrule
\end{tabular}
    \vspace{-0.5em}
\end{table}

\begin{figure}[t!]
  \centering
  \includegraphics[width = 0.9\linewidth]{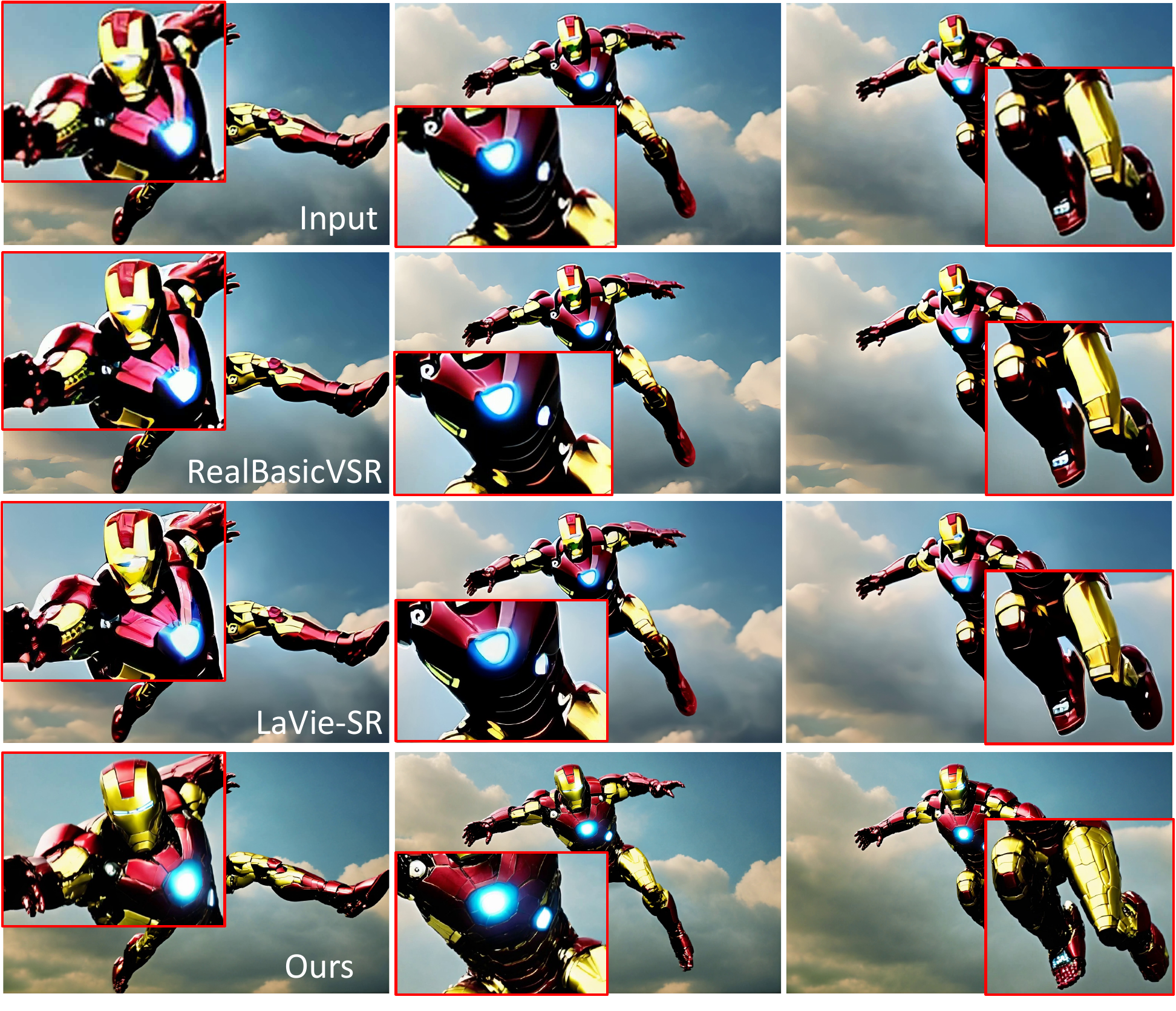} 
  \vspace{-1em}
  \caption{Visual comparison for video super-resolution ($4\times$) on AIGC2023 test dataset. Prompt: \textit{Iron Man flying in the sky}. \textbf{Zoom in for best view.}}
  \label{fig:srx4}
  \vspace{-2em}
\end{figure}

As shown in Table \ref{tab:sr_x4}, VEnhancer outperforms both generative video super-resolution method (LaVie-SR \cite{lavie}) and real-world video super-resolution method (RealBasicVSR \cite{investigating}) in all metrics, suggesting its outstanding enhancement ability for videos. 
Note that LaVie-SR surpasses RealBasicVSR in image/video quality, as LaVie-SR obtains higher scores in MUSIQ, DOVER, \textbf{Imaging Quality}, and \textbf{Aesthetic Quality}. This indicates that LaVie-SR could produce sharper results than RealBasicVSR. It is because that LaVie-SR is based on diffusion models while RealBasicVSR is trained with GAN loss. Besides, RealBasicVSR shows better performance in \textbf{Motion Smoothness} compared with LaVie-SR, since smooth results are more likely to achieve higher score in this metric. Practically, there is a trade off between
video smoothness and video quality \cite{vbench}.  
Nevertheless, VEnhancer achieves a great balance between these two aspects.

The visual comparison is presented in Fig.~\ref{fig:srx4}. The prompt is \textit{``Iron man flying in the sky"}.
The input video is already consistent with the prompt, but lacks details on the iron man suit. RealBasicVSR could
remove some noises or artifacts of the generated videos as it incorporates complex degradation for model training. However, it fails in generating realistic details but produces over-smoothed results, since its generative ability is limited.
On the other hand, the results of LaVie-SR contains more artifacts than input.
Without successfully removing artifacts, the generative super-resolution model will enlarge the existing defects. 
In contrast, VEnhancer could first remove unpleasing artifacts and refine the distorted content (\textit{e.g.,} head region), and then generate faithfuls details (\textit{e.g.,} helmet and armor) that are consistent with the text prompt. Also, the whole video is with high-quality and high-resolution, as well as smooth dynamics. 


\subsection{Comparison with Space-Time Super-Resolution Methods}
For space-time super-resolution task, we compare two state-of-the-art space-time super-resolution methods: VideoINR \cite{videoinr} and Zooming-Slow-Mo \cite{zoom} (Zoom for short).
We also consider LaVie's two-stage pipeline: LaVie-FI (frame interpolation) + LaVie-SR for more thorough comparison.

\begin{table}[htbp]
    \centering
    \small
    \vspace{-1em}
     \caption{Quantitative comparison for space-time super-resolution ($4\times$) on AIGC2023 test dataset. {\color{red}\textbf{Red}} and {\color{blue}blue} indicate the best and second best performance. The top 3 results are marked as \colorbox{mygray}{gray}.}
    \label{tab:sr_x4_fi_x4}
\begin{tabular}{c|cccccc}
\toprule
                & DOVER↑                          & MUSIQ↑                           & \makecell{Imaging\\Quality}                & \makecell{Aesthetic\\Quality}              & \makecell{Subject\\Consistency}            & \makecell{Motion\\Smoothness}              \\ \hline
LaVie-FI + LaVie-SR \cite{lavie}           & {\color[HTML]{0B5FD1} 0.8159}  & {\color[HTML]{FF0000} \textbf{53.2128}}  & {\color[HTML]{0B5FD1} 0.5299}  & 0.6566                         & \cellcolor[HTML]{F2F2F2}0.9603 & 0.9857                         \\
VideoINR  \cite{videoinr}       & \cellcolor[HTML]{F2F2F2}0.7608 & \cellcolor[HTML]{F2F2F2}34.1060 & 0.4778                         & \cellcolor[HTML]{F2F2F2}0.6624 & {\color[HTML]{0B5FD1} 0.9615}  & {\color[HTML]{FF0000} \textbf{0.9933}}  \\
Zooming Slow-Mo \cite{zoom} & 0.7328                         & 33.8470                         & \cellcolor[HTML]{F2F2F2}0.4925 & {\color[HTML]{0B5FD1} 0.6624}  & 0.9524                         & {\color[HTML]{0B5FD1} 0.9908}  \\
Ours            & {\color[HTML]{FF0000} \textbf{0.8487}}  & {\color[HTML]{0B5FD1} 50.3659}  & {\color[HTML]{FF0000} \textbf{0.5648}}  & {\color[HTML]{FF0000} \textbf{0.6665}}  & {\color[HTML]{FF0000} \textbf{0.9666}}  & \cellcolor[HTML]{F2F2F2}0.9898 \\ \bottomrule
\end{tabular}
\end{table}
\begin{figure}[htbp]
  \centering
  \includegraphics[width = 1.0\linewidth]{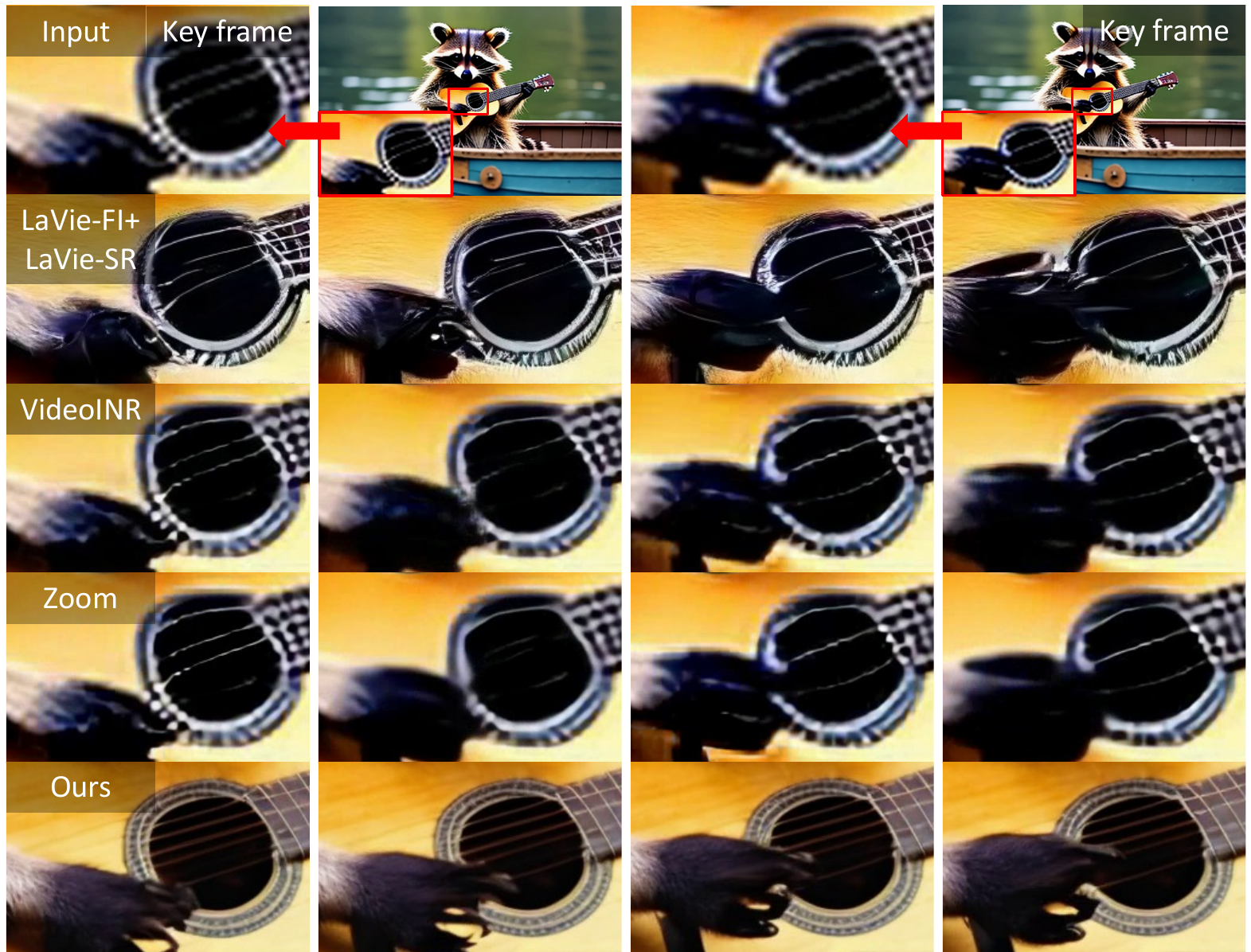}  
  \caption{Visual comparison for space-time super-resolution ($4\times$) on AIGC2023 test dataset. Prompt: \textit{A cute raccoon playing guitar in a boat on the ocean.}
\textbf{Zoom in for best view.}}
  \label{fig:srx4_fix4}
  \vspace{-2em}
\end{figure}

As shown in Table \ref{tab:sr_x4_fi_x4}, we observe that our method surpasses all baselines in DOVER, \textbf{Imaging Quality}, and \textbf{Aesthetic Quality}, showing its superior capability in generating sharp and realistic video content. Besides, it obtains highest score in \textbf{Subject Consistency}, indicating VEnhancer's ability in maintaining the subject consistent when performing joint frame interpolation and spatial super-resolution. 
We notice that state-of-the-art space-time super-resolution methods (VideoINR and Zooming Slow-Mo) stand out in \textbf{Motion Smoothness}. As both of them are optimized with reconstruction loss, the produced results are very smooth across frames. 
At a cost, they perform poorly in metrics regarding quality, such as DOVER, MUSIQ, and \textbf{Imaging Quality}.
The two-stage approach (LaVie-FI + LaVie-SR) obtains good scores in DOVER, MUSIQ and \textbf{Imaging Quality}, demonstrating DM-based methods' advantage in generation. However, its performance in video stability and subject consistency is unsatisfactory due to its inferior capability in video refinement. 

The visual comparison is illustrated in Fig.~\ref{fig:srx4_fix4}.
The first and third columns present the low-resolution key frames.
Note that the input frames are not consistent especially in the region of guitar strings. The two-stage DM-based approach, LaVie-FI + LaVie-SR \cite{lavie}, can produce very sharp results for all frames (key and predicted ones). However, it generates messy contents which are not semantically aligned with prompt. Moreover, the generated details are changing across time, indicating severe flickering.
For reconstruction-based methods (VideoINR \cite{videoinr} and Zoom \cite{zoom}), the produced results are similar: lacking details and fail in improving the consistency of the original input frames. 
On the contrary, VEnhancer is not only able to achieve unified space-time super-resolution, but can also improve the temporal consistency of the generated videos by refinement (\textit{i.e.,} guitar strings and raccoon hands plucking the strings). 
\label{sec:ST_fig}

\subsection{Evaluation on Improving Video Generation}
In this section, we evaluate VEnhancer's ability in improving the existing state-of-the-art T2V method (\textit{i.e.,} VideoCrafter2 \cite{videocrafter2}). The baselines includes open-source T2V methods: VideoCrafter-2 (VC-2 for short), Show-1 \cite{show1}, Lavie \cite{lavie}, Open-Sora \footnote{https://github.com/hpcaitech/Open-Sora}, as well as professional video generation products, Pika and Gen-2. All dimensions of VBench are considered for evaluation. 

The quantitative results are organized in Table \ref{tab:vbench}. 
In general, VideoCrafter-2 is already the best in overall \textit{Semantic} compared with other baselines, demonstrating its superiority in generating high-fidelity contents to the VBench's prompt suite. Regarding the overall \textit{Quality}, it lags behind Pika and Gen-2. 
But with VEnhancer, VideoCrafter-2 is able to achieve the highest scores in both overall \textit{Quality} and \textit{Semantic}. This indicates that VEnhancer can improve the semantic content and video quality at the same time, showing a powerful unified enhancement (\textit{i.e.,} refinement and space-time super-resolution) ability.

\begin{figure}[htbp]
  \centering
  \includegraphics[width = 1.0\linewidth]{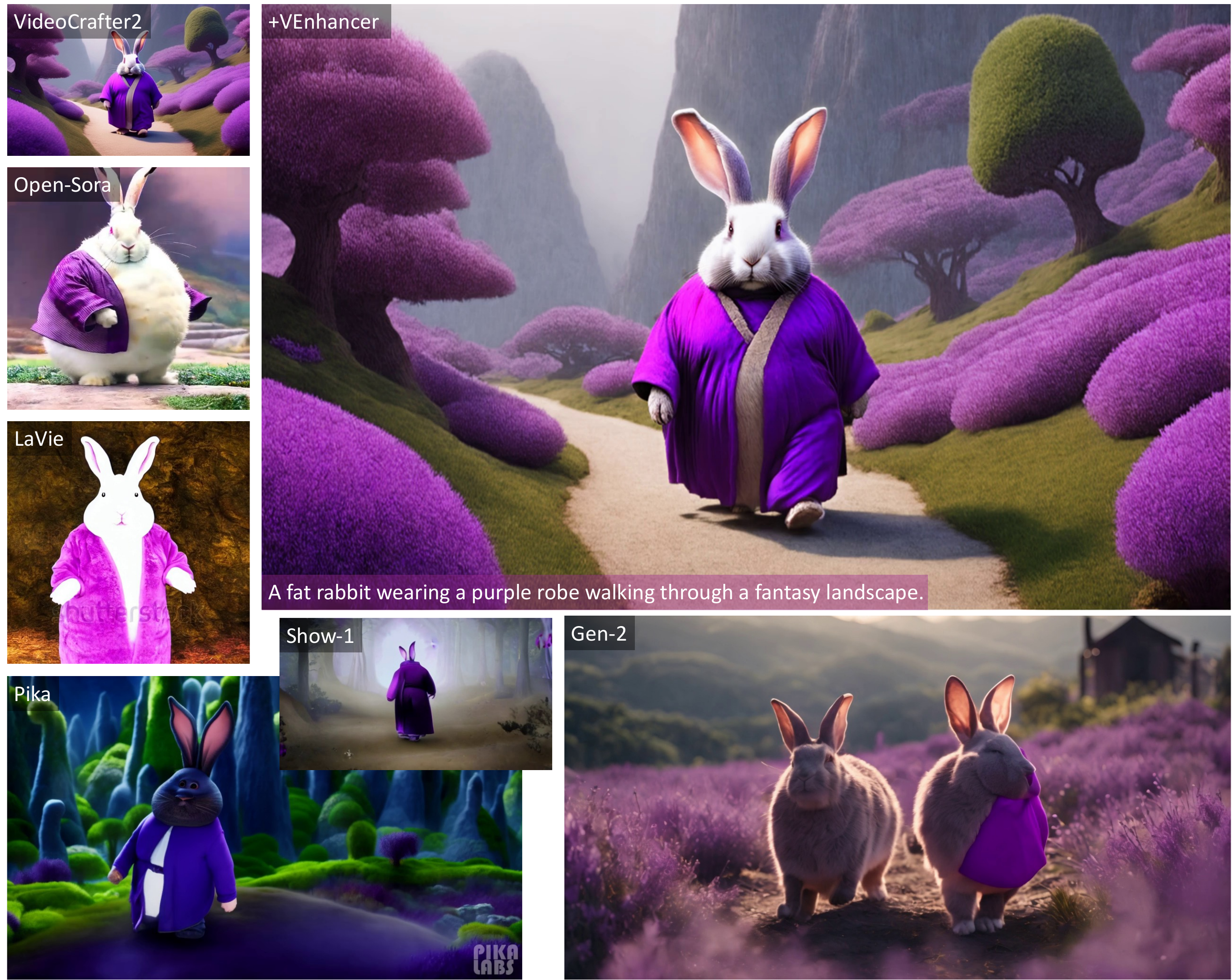}  
  \caption{Visual comparison of screenshots obtained from VideoCrafter-2+VEnhancer and other T2V models. The layout is arranged based on their original resolutions.  \textbf{Zoom in for best view.}}
  \label{fig:vbench}
\end{figure}

Regarding \textit{Quality} aspect, we note that Pika and Gen-2 could obtain good scores in various dimensions, but perform very poorly in \textbf{dynamic degree}. This dimension assesses whether a video has large motions, and higher score corresponds to larger motion. It is suggested that Gen-2 and Pika 
obtain good temporal consistency by sacrificing versertile motions. 
This demonstrates the advantage of adopting a two-stage pipeline: the first T2V model focuses on generating semantic content and motions with good fidelity to the prompts, while the following enhancement model can improve the semantic in low-level and image quality, as well as temporal consistency. We also provide visual comparison in Fig.~\ref{fig:vbench}. It is obvious that VideoCrafter-2 yields the best video content but with low-resolution. And VEnhancer significantly improves its quality (with 2k resolution). Gen-2 and Pika could produce results with around 1k resolution, but their generated contents have some apparent flaws (weird robe in Gen-2 and unnatural rabbit’face in Pika). Open-Sora and Lavie both generate videos with square size, leading to incomplete generation for backgrounds.

\begingroup
\renewcommand{\arraystretch}{1.25}
\begin{table}[htbp]
    \centering
     \caption{VBench Evaluation Results per Dimension. This table compares the performance of four open-source video generation models (LaVie \cite{lavie}, Show-1 \cite{show1}, Open-Sora, VideoCrafter-2 \cite{videocrafter2}) and two professional video generation products (Pika and Gen-2) across
each of the 16 VBench dimensions regarding two aspects (\textit{Quality} and \textit{Semantic}). A higher score indicates relatively better performance for a particular dimension. {\color{red}\textbf{Red}} and {\color{blue}blue} indicate the best and second best performance. The top 3 results are marked as \colorbox{mygray}{gray}.}
    \resizebox{\linewidth}{!}{
\begin{tabular}{c|c|ccccccc}
\hline
\textbf{}                           & \textbf{Dimensions}             & \textbf{Show-1} \cite{show1}                & \textbf{LaVie} \cite{lavie}                  & \textbf{Open-Sora}              & \textbf{Pika}                                          & \textbf{Gen-2}                  & \textbf{VC-2} \cite{videocrafter2}                                         & \textbf{VC-2+VEnhancer}                                \\ \hline
                                    & \textbf{Subject Consistency}    & 95.53\%                         & 91.41\%                         & 92.09\%                         & 96.76\%                                                & \cellcolor[HTML]{F2F2F2}97.61\% & \cellcolor[HTML]{F2F2F2}96.85\%                        & \cellcolor[HTML]{F2F2F2}97.17\%                        \\
                                    & \textbf{Background Consistency} & 98.02\%                         & 97.47\%                         & 97.39\%                         & \cellcolor[HTML]{F2F2F2}98.95\%                        & 97.61\%                         & \cellcolor[HTML]{F2F2F2}98.22\%                        & \cellcolor[HTML]{F2F2F2}98.54\%                        \\
                                    & \textbf{Temporal Flickering}    & \cellcolor[HTML]{F2F2F2}99.12\% & 98.30\%                         & 98.41\%                         & \cellcolor[HTML]{F2F2F2}99.77\%                        & \cellcolor[HTML]{F2F2F2}99.56\% & 98.41\%                                                & 98.46\%                                                \\
                                    & \textbf{Motion Smoothness}      & \cellcolor[HTML]{F2F2F2}98.24\% & 96.38\%                         & 95.61\%                         & \cellcolor[HTML]{F2F2F2}99.51\%                        & \cellcolor[HTML]{F2F2F2}99.58\% & 97.73\%                                                & 97.75\%                                                \\
                                    & \textbf{Aesthetic Quality}      & 57.35\%                         & 54.94\%                         & 57.76\%                         & \cellcolor[HTML]{F2F2F2}63.15\%                        & \cellcolor[HTML]{F2F2F2}66.96\% & 63.13\%                                                & \cellcolor[HTML]{F2F2F2}65.89\%                        \\
                                    & \textbf{Dynamic Degree}         & \cellcolor[HTML]{F2F2F2}44.44\% & \cellcolor[HTML]{F2F2F2}49.72\% & \cellcolor[HTML]{F2F2F2}48.61\% & 37.22\%                                                & 18.89\%                         & 42.50\%                                                & 42.50\%                                                \\
\multirow{-7}{*}{\textbf{Quality}}  & \textbf{Imaging Quality}        & 58.66\%                         & 61.90\%                         & 61.51\%                         & 62.33\%                                                & \cellcolor[HTML]{F2F2F2}67.42\% & \cellcolor[HTML]{F2F2F2}67.22\%                        & \cellcolor[HTML]{F2F2F2}70.45\%                        \\ \hline
                                    & \textbf{Object Class}           & \cellcolor[HTML]{F2F2F2}93.07\% & 91.82\%                         & 74.98\%                         & 87.45\%                                                & 90.92\%                         & \cellcolor[HTML]{F2F2F2}92.55\%                        & \cellcolor[HTML]{F2F2F2}93.39\%                        \\
                                    & \textbf{Multiple Objects}       & 45.47\%                         & 33.32\%                         & 33.64\%                         & \cellcolor[HTML]{F2F2F2}46.69\%                        & \cellcolor[HTML]{F2F2F2}55.47\% & 40.66\%                                                & \cellcolor[HTML]{F2F2F2}49.83\%                        \\
                                    & \textbf{Human Action}           & \cellcolor[HTML]{F2F2F2}95.60\% & \cellcolor[HTML]{F2F2F2}96.80\% & 85.00\%                         & 88.00\%                                                & 89.20\%                         & \cellcolor[HTML]{F2F2F2}95.00\%                        & \cellcolor[HTML]{F2F2F2}95.00\%                        \\
                                    & \textbf{Color}                  & 86.35\%                         & 86.39\%                         & 78.15\%                         & 85.31\%                                                & \cellcolor[HTML]{F2F2F2}89.49\% & \cellcolor[HTML]{F2F2F2}92.92\%                        & \cellcolor[HTML]{F2F2F2}94.41\%                        \\
                                    & \textbf{Spatial Relationship}   & 53.50\%                         & 34.09\%                         & 43.95\%                         & \cellcolor[HTML]{F2F2F2}65.65\%                        & \cellcolor[HTML]{F2F2F2}66.91\% & 35.86\%                                                & \cellcolor[HTML]{F2F2F2}64.88\%                        \\
                                    & \textbf{Scene}                  & 47.03\%                         & \cellcolor[HTML]{F2F2F2}52.69\% & 37.33\%                         & 44.80\%                                                & 48.91\%                         & \cellcolor[HTML]{F2F2F2}55.29\%                        & \cellcolor[HTML]{F2F2F2}51.82\%                        \\
                                    & \textbf{Appearance Style}       & 23.06\%                         & \cellcolor[HTML]{F2F2F2}23.56\% & 21.58\%                         & 21.89\%                                                & 19.34\%                         & \cellcolor[HTML]{F2F2F2}25.13\%                        & \cellcolor[HTML]{F2F2F2}24.32\%                        \\
                                    & \textbf{Temporal Style}         & 25.28\%                         & \cellcolor[HTML]{F2F2F2}25.93\% & \cellcolor[HTML]{F2F2F2}25.46\% & 24.44\%                                                & 24.12\%                         & \cellcolor[HTML]{F2F2F2}25.84\%                        & 25.17\%                                                \\
\multirow{-9}{*}{\textbf{Semantic}} & \textbf{Overall Consistency}    & \cellcolor[HTML]{F2F2F2}27.46\% & 26.41\%                         & 26.18\%                         & 25.47\%                                                & 26.17\%                         & \cellcolor[HTML]{F2F2F2}28.23\%                        & \cellcolor[HTML]{F2F2F2}27.57\%                        \\ \hline
                                    & \textbf{Quality}                & 80.42\%                         & 78.78\%                         & 78.82\%                         & \cellcolor[HTML]{F2F2F2}{\color[HTML]{0B5FD1} 82.68\%} & \cellcolor[HTML]{F2F2F2}82.46\% & 82.20\%                                                & \cellcolor[HTML]{F2F2F2}{\color[HTML]{FF0000} \textbf{83.28\%}} \\
\multirow{-2}{*}{\textbf{Overall}}  & \textbf{Semantic}               & 72.98\%                         & 70.31\%                         & 64.28\%                         & 71.26\%                                                & \cellcolor[HTML]{F2F2F2}73.03\% & \cellcolor[HTML]{F2F2F2}{\color[HTML]{0B5FD1} 73.42\%} & \cellcolor[HTML]{F2F2F2}{\color[HTML]{FF0000} \textbf{76.73\%}} \\ \hline
\end{tabular}
}
    \label{tab:vbench}
\end{table}
\endgroup


\subsection{Ablation Studies}
Please refer to Appendix \ref{sec:Ablation} for arbitrary up-sampling scales for space/time super-resolution, as well as different noise levels in noise augmentation.

\section{Conclusion and Limitation.}
In this work, we propose a generative space-time enhancement method, VEnhancer, for video generation. It can achieve spatial super-resolution, temporal super-resolution and video refinement in a unified framework. We base on a pretrained video diffusion model and build a trainable video ControlNet for effective condition injection. Space-time data augmentation and video-aware conditioning are proposed to train video ControlNet in an end-to-end manner. Extensive experiments have demonstrated the superiority over state-of-the-art video super-resolution and space-time
super-resolution methods in enhancing AI-generated videos. Moreover, VEnhancer is able to improve the results of existing state-of-the-art T2V method, lifting VideoCrater-2's ranking to top one. However, our work has several limitations. First, as it is based on diffusion models, the inference takes more time than one-step methods, for instance, reconstruction-based space-time super-resolution methods. Second, it may face challenges in handling AI-generated long videos, since the long-term (over 10s) consistency has not been addressed in this work.

\newpage

{\small
\bibliographystyle{plain}
\bibliography{reference}

\begin{thebibliography}{10}

\bibitem{webvid}
Max Bain, Arsha Nagrani, G{\"u}l Varol, and Andrew Zisserman.
\newblock Frozen in time: A joint video and image encoder for end-to-end retrieval.
\newblock In {\em IEEE International Conference on Computer Vision}, 2021.

\bibitem{svd}
Andreas Blattmann, Tim Dockhorn, Sumith Kulal, Daniel Mendelevitch, Maciej Kilian, Dominik Lorenz, Yam Levi, Zion English, Vikram Voleti, Adam Letts, et~al.
\newblock Stable video diffusion: Scaling latent video diffusion models to large datasets.
\newblock {\em arXiv preprint arXiv:2311.15127}, 2023.

\bibitem{align}
Andreas Blattmann, Robin Rombach, Huan Ling, Tim Dockhorn, Seung~Wook Kim, Sanja Fidler, and Karsten Kreis.
\newblock Align your latents: High-resolution video synthesis with latent diffusion models.
\newblock In {\em Proceedings of the IEEE/CVF Conference on Computer Vision and Pattern Recognition}, pages 22563--22575, 2023.

\bibitem{vsr_transformer}
Jiezhang Cao, Yawei Li, Kai Zhang, and Luc Van~Gool.
\newblock Video super-resolution transformer.
\newblock {\em arXiv preprint arXiv:2106.06847}, 2021.

\bibitem{basicvsr}
Kelvin~CK Chan, Xintao Wang, Ke~Yu, Chao Dong, and Chen~Change Loy.
\newblock Basicvsr: The search for essential components in video super-resolution and beyond.
\newblock In {\em Proceedings of the IEEE/CVF conference on computer vision and pattern recognition}, pages 4947--4956, 2021.

\bibitem{basicvsr++}
Kelvin~CK Chan, Shangchen Zhou, Xiangyu Xu, and Chen~Change Loy.
\newblock Basicvsr++: Improving video super-resolution with enhanced propagation and alignment.
\newblock In {\em Proceedings of the IEEE/CVF conference on computer vision and pattern recognition}, pages 5972--5981, 2022.

\bibitem{investigating}
Kelvin~CK Chan, Shangchen Zhou, Xiangyu Xu, and Chen~Change Loy.
\newblock Investigating tradeoffs in real-world video super-resolution.
\newblock In {\em Proceedings of the IEEE/CVF Conference on Computer Vision and Pattern Recognition}, pages 5962--5971, 2022.

\bibitem{videocrafter1}
Haoxin Chen, Menghan Xia, Yingqing He, Yong Zhang, Xiaodong Cun, Shaoshu Yang, Jinbo Xing, Yaofang Liu, Qifeng Chen, Xintao Wang, et~al.
\newblock Videocrafter1: Open diffusion models for high-quality video generation.
\newblock {\em arXiv preprint arXiv:2310.19512}, 2023.

\bibitem{videocrafter2}
Haoxin Chen, Yong Zhang, Xiaodong Cun, Menghan Xia, Xintao Wang, Chao Weng, and Ying Shan.
\newblock Videocrafter2: Overcoming data limitations for high-quality video diffusion models.
\newblock {\em arXiv preprint arXiv:2401.09047}, 2024.

\bibitem{pixart}
Junsong Chen, Jincheng Yu, Chongjian Ge, Lewei Yao, Enze Xie, Yue Wu, Zhongdao Wang, James Kwok, Ping Luo, Huchuan Lu, et~al.
\newblock Pixart-$\alpha$: Fast training of diffusion transformer for photorealistic text-to-image synthesis.
\newblock {\em arXiv preprint arXiv:2310.00426}, 2023.

\bibitem{gentron}
Shoufa Chen, Mengmeng Xu, Jiawei Ren, Yuren Cong, Sen He, Yanping Xie, Animesh Sinha, Ping Luo, Tao Xiang, and Juan-Manuel Perez-Rua.
\newblock Gentron: Delving deep into diffusion transformers for image and video generation.
\newblock {\em arXiv preprint arXiv:2312.04557}, 2023.

\bibitem{videoinr}
Zeyuan Chen, Yinbo Chen, Jingwen Liu, Xingqian Xu, Vidit Goel, Zhangyang Wang, Humphrey Shi, and Xiaolong Wang.
\newblock Videoinr: Learning video implicit neural representation for continuous space-time super-resolution.
\newblock In {\em Proceedings of the IEEE/CVF Conference on Computer Vision and Pattern Recognition}, pages 2047--2057, 2022.

\bibitem{silu}
Stefan Elfwing, Eiji Uchibe, and Kenji Doya.
\newblock Sigmoid-weighted linear units for neural network function approximation in reinforcement learning.
\newblock {\em Neural networks}, 107:3--11, 2018.

\bibitem{lumina}
Peng Gao, Le~Zhuo, Ziyi Lin, Chris Liu, Junsong Chen, Ruoyi Du, Enze Xie, Xu~Luo, Longtian Qiu, Yuhang Zhang, et~al.
\newblock Lumina-t2x: Transforming text into any modality, resolution, and duration via flow-based large diffusion transformers.
\newblock {\em arXiv preprint arXiv:2405.05945}, 2024.

\bibitem{animatediff}
Yuwei Guo, Ceyuan Yang, Anyi Rao, Yaohui Wang, Yu~Qiao, Dahua Lin, and Bo~Dai.
\newblock Animatediff: Animate your personalized text-to-image diffusion models without specific tuning.
\newblock {\em arXiv preprint arXiv:2307.04725}, 2023.

\bibitem{gupta2023photorealistic}
Agrim Gupta, Lijun Yu, Kihyuk Sohn, Xiuye Gu, Meera Hahn, Li~Fei-Fei, Irfan Essa, Lu~Jiang, and Jos{\'e} Lezama.
\newblock Photorealistic video generation with diffusion models.
\newblock {\em arXiv preprint arXiv:2312.06662}, 2023.

\bibitem{starnet}
Muhammad Haris, Greg Shakhnarovich, and Norimichi Ukita.
\newblock Space-time-aware multi-resolution video enhancement.
\newblock In {\em Proceedings of the IEEE/CVF conference on computer vision and pattern recognition}, pages 2859--2868, 2020.

\bibitem{resnet}
Kaiming He, Xiangyu Zhang, Shaoqing Ren, and Jian Sun.
\newblock Deep residual learning for image recognition.
\newblock In {\em Proceedings of the IEEE conference on computer vision and pattern recognition}, pages 770--778, 2016.

\bibitem{imagenvideo}
Jonathan Ho, William Chan, Chitwan Saharia, Jay Whang, Ruiqi Gao, Alexey Gritsenko, Diederik~P Kingma, Ben Poole, Mohammad Norouzi, David~J Fleet, et~al.
\newblock Imagen video: High definition video generation with diffusion models.
\newblock {\em arXiv preprint arXiv:2210.02303}, 2022.

\bibitem{ddpm}
Jonathan Ho, Ajay Jain, and Pieter Abbeel.
\newblock Denoising diffusion probabilistic models.
\newblock {\em Advances in neural information processing systems}, 33:6840--6851, 2020.

\bibitem{cascaded}
Jonathan Ho, Chitwan Saharia, William Chan, David~J Fleet, Mohammad Norouzi, and Tim Salimans.
\newblock Cascaded diffusion models for high fidelity image generation.
\newblock {\em Journal of Machine Learning Research}, 23(47):1--33, 2022.

\bibitem{cfg}
Jonathan Ho and Tim Salimans.
\newblock Classifier-free diffusion guidance.
\newblock {\em arXiv preprint arXiv:2207.12598}, 2022.

\bibitem{vbench}
Ziqi Huang, Yinan He, Jiashuo Yu, Fan Zhang, Chenyang Si, Yuming Jiang, Yuanhan Zhang, Tianxing Wu, Qingyang Jin, Nattapol Chanpaisit, et~al.
\newblock Vbench: Comprehensive benchmark suite for video generative models.
\newblock {\em arXiv preprint arXiv:2311.17982}, 2023.

\bibitem{vsr_recurrent}
Takashi Isobe, Xu~Jia, Shuhang Gu, Songjiang Li, Shengjin Wang, and Qi~Tian.
\newblock Video super-resolution with recurrent structure-detail network.
\newblock In {\em Computer Vision--ECCV 2020: 16th European Conference, Glasgow, UK, August 23--28, 2020, Proceedings, Part XII 16}, pages 645--660. Springer, 2020.

\bibitem{vsr_tga}
Takashi Isobe, Songjiang Li, Xu~Jia, Shanxin Yuan, Gregory Slabaugh, Chunjing Xu, Ya-Li Li, Shengjin Wang, and Qi~Tian.
\newblock Video super-resolution with temporal group attention.
\newblock In {\em Proceedings of the IEEE/CVF conference on computer vision and pattern recognition}, pages 8008--8017, 2020.

\bibitem{revisiting}
Takashi Isobe, Fang Zhu, Xu~Jia, and Shengjin Wang.
\newblock Revisiting temporal modeling for video super-resolution.
\newblock {\em arXiv preprint arXiv:2008.05765}, 2020.

\bibitem{musiq}
Junjie Ke, Qifei Wang, Yilin Wang, Peyman Milanfar, and Feng Yang.
\newblock Musiq: Multi-scale image quality transformer.
\newblock In {\em Proceedings of the IEEE/CVF International Conference on Computer Vision}, pages 5148--5157, 2021.

\bibitem{fisr}
Soo~Ye Kim, Jihyong Oh, and Munchurl Kim.
\newblock Fisr: Deep joint frame interpolation and super-resolution with a multi-scale temporal loss.
\newblock In {\em Proceedings of the AAAI Conference on Artificial Intelligence}, volume~34, pages 11278--11286, 2020.

\bibitem{vrt}
Jingyun Liang, Jiezhang Cao, Yuchen Fan, Kai Zhang, Rakesh Ranjan, Yawei Li, Radu Timofte, and Luc Van~Gool.
\newblock Vrt: A video restoration transformer.
\newblock {\em IEEE Transactions on Image Processing}, 2024.

\bibitem{recurrent}
Jingyun Liang, Yuchen Fan, Xiaoyu Xiang, Rakesh Ranjan, Eddy Ilg, Simon Green, Jiezhang Cao, Kai Zhang, Radu Timofte, and Luc~V Gool.
\newblock Recurrent video restoration transformer with guided deformable attention.
\newblock {\em Advances in Neural Information Processing Systems}, 35:378--393, 2022.

\bibitem{diffbir}
Xinqi Lin, Jingwen He, Ziyan Chen, Zhaoyang Lyu, Bo~Dai, Fanghua Yu, Wanli Ouyang, Yu~Qiao, and Chao Dong.
\newblock Diffbir: Towards blind image restoration with generative diffusion prior, 2024.

\bibitem{adamw}
Ilya Loshchilov and Frank Hutter.
\newblock Decoupled weight decay regularization.
\newblock {\em arXiv preprint arXiv:1711.05101}, 2017.

\bibitem{sdedit}
Chenlin Meng, Yutong He, Yang Song, Jiaming Song, Jiajun Wu, Jun-Yan Zhu, and Stefano Ermon.
\newblock Sdedit: Guided image synthesis and editing with stochastic differential equations.
\newblock {\em arXiv preprint arXiv:2108.01073}, 2021.

\bibitem{sdxl}
Dustin Podell, Zion English, Kyle Lacey, Andreas Blattmann, Tim Dockhorn, Jonas M{\"u}ller, Joe Penna, and Robin Rombach.
\newblock Sdxl: Improving latent diffusion models for high-resolution image synthesis.
\newblock {\em arXiv preprint arXiv:2307.01952}, 2023.

\bibitem{sd}
Robin Rombach, Andreas Blattmann, Dominik Lorenz, Patrick Esser, and Bj{\"o}rn Ommer.
\newblock High-resolution image synthesis with latent diffusion models.
\newblock In {\em Proceedings of the IEEE/CVF conference on computer vision and pattern recognition}, pages 10684--10695, 2022.

\bibitem{vprediction}
Tim Salimans and Jonathan Ho.
\newblock Progressive distillation for fast sampling of diffusion models.
\newblock {\em arXiv preprint arXiv:2202.00512}, 2022.

\bibitem{ddim}
Jiaming Song, Chenlin Meng, and Stefano Ermon.
\newblock Denoising diffusion implicit models.
\newblock {\em arXiv preprint arXiv:2010.02502}, 2020.

\bibitem{attention}
Ashish Vaswani, Noam Shazeer, Niki Parmar, Jakob Uszkoreit, Llion Jones, Aidan~N Gomez, {\L}ukasz Kaiser, and Illia Polosukhin.
\newblock Attention is all you need.
\newblock {\em Advances in neural information processing systems}, 30, 2017.

\bibitem{modelscope}
Jiuniu Wang, Hangjie Yuan, Dayou Chen, Yingya Zhang, Xiang Wang, and Shiwei Zhang.
\newblock Modelscope text-to-video technical report.
\newblock {\em arXiv preprint arXiv:2308.06571}, 2023.

\bibitem{videocomposer}
Xiang Wang, Hangjie Yuan, Shiwei Zhang, Dayou Chen, Jiuniu Wang, Yingya Zhang, Yujun Shen, Deli Zhao, and Jingren Zhou.
\newblock Videocomposer: Compositional video synthesis with motion controllability.
\newblock {\em Advances in Neural Information Processing Systems}, 36, 2024.

\bibitem{edvr}
Xintao Wang, Kelvin~CK Chan, Ke~Yu, Chao Dong, and Chen Change~Loy.
\newblock Edvr: Video restoration with enhanced deformable convolutional networks.
\newblock In {\em Proceedings of the IEEE/CVF conference on computer vision and pattern recognition workshops}, pages 0--0, 2019.

\bibitem{lavie}
Yaohui Wang, Xinyuan Chen, Xin Ma, Shangchen Zhou, Ziqi Huang, Yi~Wang, Ceyuan Yang, Yinan He, Jiashuo Yu, Peiqing Yang, et~al.
\newblock Lavie: High-quality video generation with cascaded latent diffusion models.
\newblock {\em arXiv preprint arXiv:2309.15103}, 2023.

\bibitem{dover}
Haoning Wu, Erli Zhang, Liang Liao, Chaofeng Chen, Jingwen~Hou Hou, Annan Wang, Wenxiu~Sun Sun, Qiong Yan, and Weisi Lin.
\newblock Exploring video quality assessment on user generated contents from aesthetic and technical perspectives.
\newblock In {\em International Conference on Computer Vision (ICCV)}, 2023.

\bibitem{tuneavideo}
Jay~Zhangjie Wu, Yixiao Ge, Xintao Wang, Stan~Weixian Lei, Yuchao Gu, Yufei Shi, Wynne Hsu, Ying Shan, Xiaohu Qie, and Mike~Zheng Shou.
\newblock Tune-a-video: One-shot tuning of image diffusion models for text-to-video generation.
\newblock In {\em Proceedings of the IEEE/CVF International Conference on Computer Vision}, pages 7623--7633, 2023.

\bibitem{zoom}
Xiaoyu Xiang, Yapeng Tian, Yulun Zhang, Yun Fu, Jan~P Allebach, and Chenliang Xu.
\newblock Zooming slow-mo: Fast and accurate one-stage space-time video super-resolution.
\newblock In {\em Proceedings of the IEEE/CVF conference on computer vision and pattern recognition}, pages 3370--3379, 2020.

\bibitem{mitigating}
Liangbin Xie, Xintao Wang, Shuwei Shi, Jinjin Gu, Chao Dong, and Ying Shan.
\newblock Mitigating artifacts in real-world video super-resolution models.
\newblock In {\em Proceedings of the AAAI Conference on Artificial Intelligence}, volume~37, pages 2956--2964, 2023.

\bibitem{video_flow}
Tianfan Xue, Baian Chen, Jiajun Wu, Donglai Wei, and William~T Freeman.
\newblock Video enhancement with task-oriented flow.
\newblock {\em International Journal of Computer Vision}, 127:1106--1125, 2019.

\bibitem{show1}
David~Junhao Zhang, Jay~Zhangjie Wu, Jia-Wei Liu, Rui Zhao, Lingmin Ran, Yuchao Gu, Difei Gao, and Mike~Zheng Shou.
\newblock Show-1: Marrying pixel and latent diffusion models for text-to-video generation.
\newblock {\em arXiv preprint arXiv:2309.15818}, 2023.

\bibitem{controlnet}
Lvmin Zhang, Anyi Rao, and Maneesh Agrawala.
\newblock Adding conditional control to text-to-image diffusion models.
\newblock In {\em Proceedings of the IEEE/CVF International Conference on Computer Vision}, pages 3836--3847, 2023.

\bibitem{i2vgenxl}
Shiwei Zhang, Jiayu Wang, Yingya Zhang, Kang Zhao, Hangjie Yuan, Zhiwu Qin, Xiang Wang, Deli Zhao, and Jingren Zhou.
\newblock I2vgen-xl: High-quality image-to-video synthesis via cascaded diffusion models.
\newblock {\em arXiv preprint arXiv:2311.04145}, 2023.

\bibitem{upscale}
Shangchen Zhou, Peiqing Yang, Jianyi Wang, Yihang Luo, and Chen~Change Loy.
\newblock Upscale-a-video: Temporal-consistent diffusion model for real-world video super-resolution.
\newblock {\em arXiv preprint arXiv:2312.06640}, 2023.

\end{thebibliography}
}

\newpage
\appendix




\section{Ablation Studies}
\label{sec:Ablation}
\subsection{The Effectiveness of Noise Augmentation.}
\begin{figure}[htbp]
  \centering
  \includegraphics[width = 1.0\linewidth]{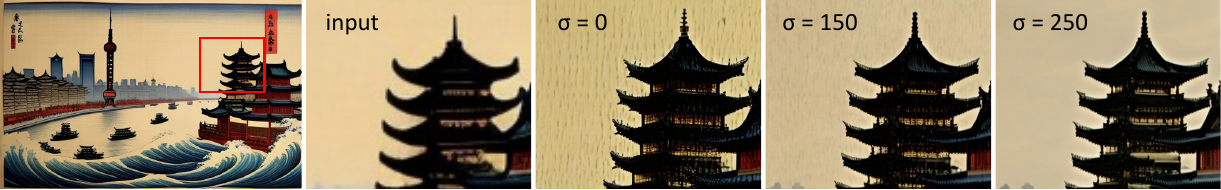}  
  \caption{Visual comparison of setting different noise levels in noise augmentation during testing.}
  \label{fig:noise_level}
\end{figure}

During training, the noise level regarding noise augmentation is randomly sampled within a predefined range. While during inference, one can change the noise level to achieve refinement with different strengths. In general, higher noise corresponds to stronger refinement and regeneration.
We present the visual comparison among different noise levels in Figure \ref{fig:noise_level}. The first frame of one AI-generated video is presented in the left. It is of low-resolution and lacks details. Also, the original video has very obvious flickering. If we set $\sigma=0$, VEnhancer will generate unpleasing noises in the background. As there is domain mismatch between the training data and testing data, the enhancement fails in handling unseen and challenging scenarios. Fortunately, we can mitigate this by adding noise in the condition latents for corrupting the noisy and unknown low-level details. As we increase the noise level, the artifacts are gradually vanishing. When $\sigma=250$, the result is noise-clean, and has abundant semantic details. 

\subsection{Arbitrary Up-sampling Scales for Spatial Super-Resolution.}
\begin{figure}[htbp]
  \centering
  \includegraphics[width = 1.0\linewidth]{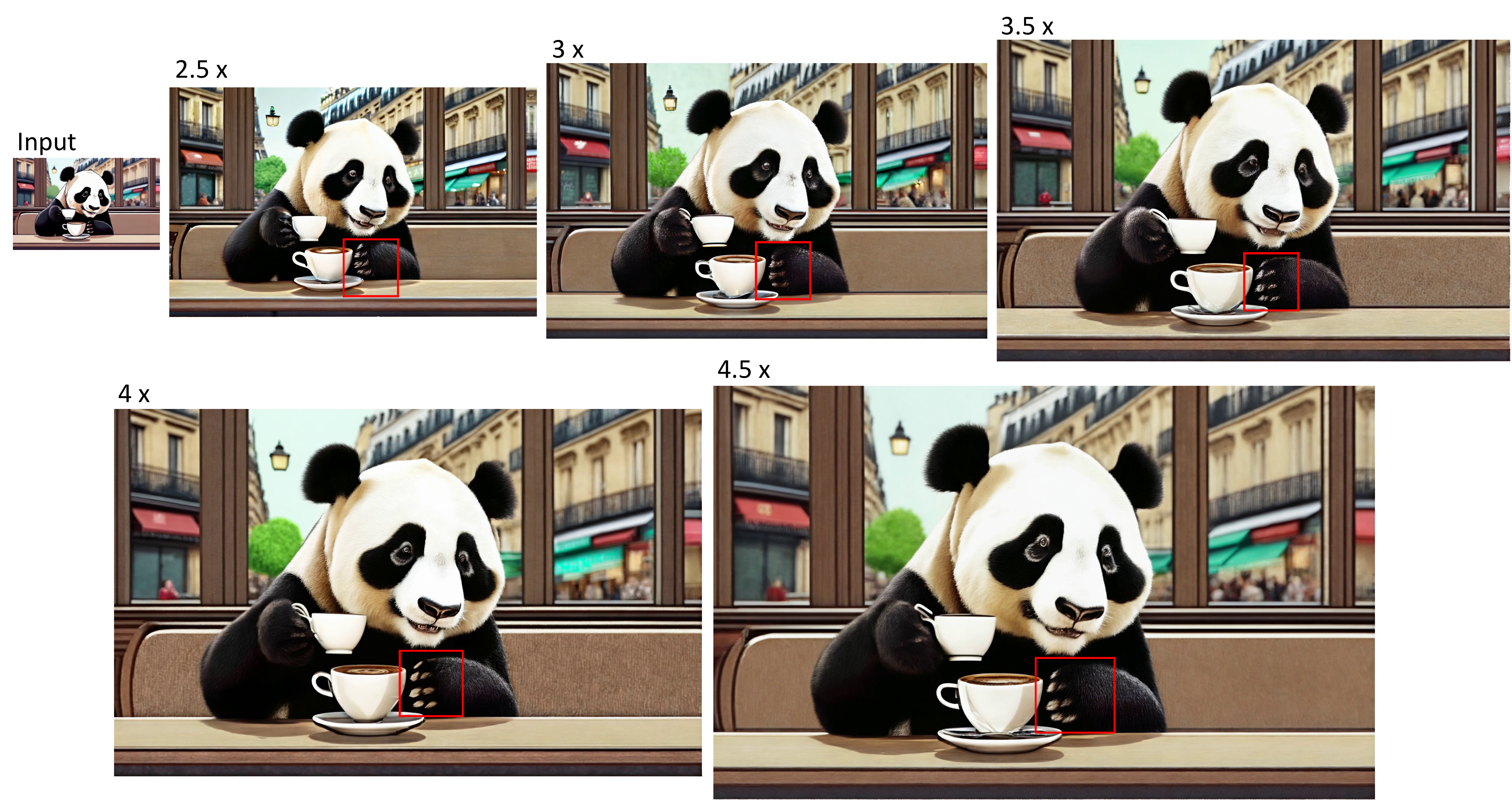}  
  \caption{Visual results of different up-sampling scales ($2.5\times$, $3\times$, $3.5\times$, $4\times$, and $4.5\times$) for spatial super-resolution during testing.}
  \label{fig:multi_scale}
\end{figure}

In this section, we show that VEnhancer is able to up-sample videos with arbitrary scales. 
From Figure \ref{fig:multi_scale}, we observe that VEnhancer could produce satisfactory results on different scales ($2.5\times$, $3\times$, $3.5\times$, $4\times$, and $4.5\times$), suggesting its flexibility and generalization in adapting to different tasks.
In particular, given one frame of the generated video ($312\times512$), VEnhancer could improve the generated details when the up-sampling scale grows up. When $s=2.5\sim3.5$, the panda's hand is less realistic. But it becomes better when $s=4$ or $s=4.5$. It is also noticed that the panda' fur is becoming more realistic as $s$ grows. 

\subsection{Arbitrary Up-sampling Scales for Temporal Super-Resolution.}

\begin{figure}[htbp]
  \centering
  \includegraphics[width = 1.0\linewidth]{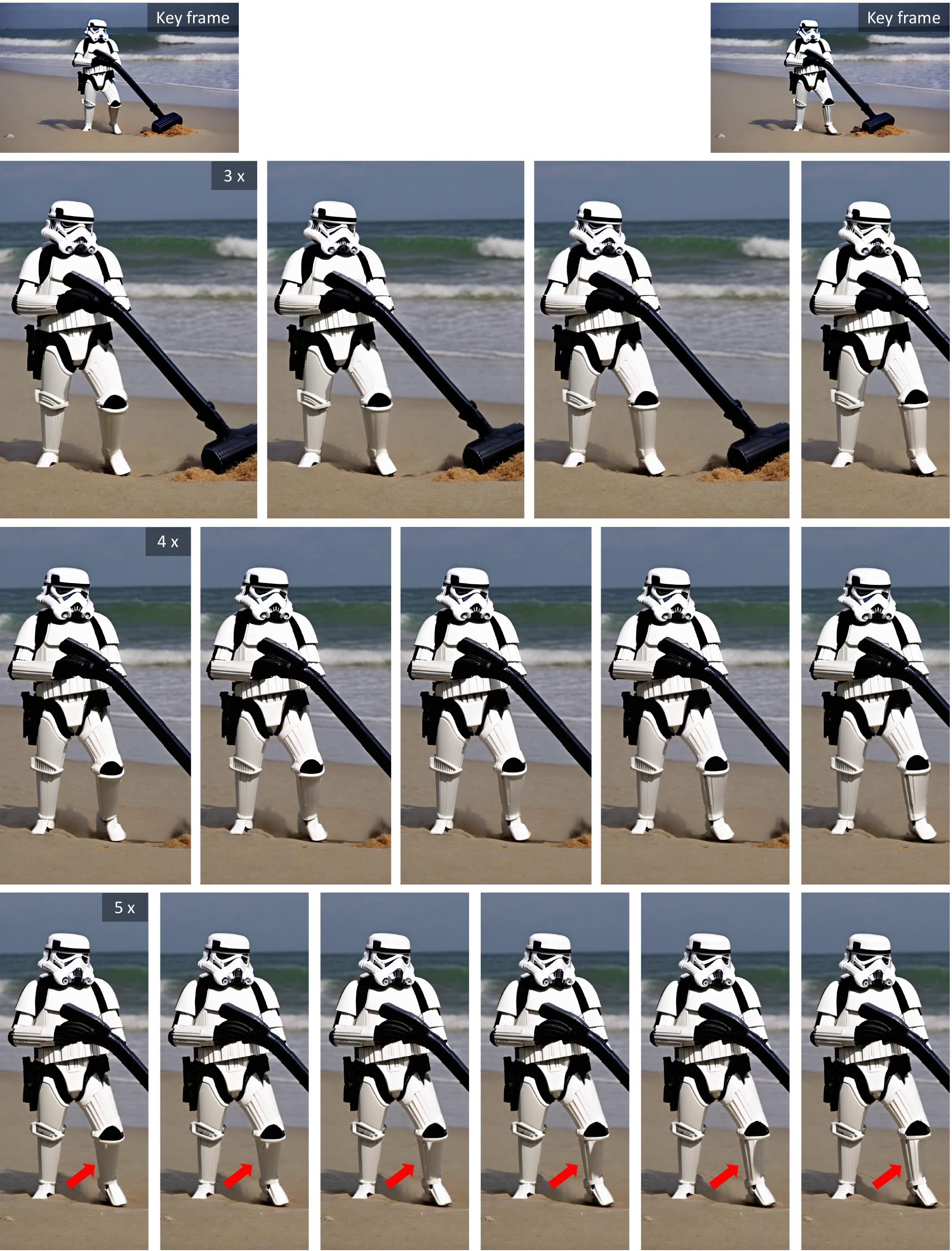}  
  \caption{Visual results of different up-sampling scales ($3\times$, $4\times$, and $5\times$) for temporal super-resolution during testing.}
  \label{fig:multi_interp}
\end{figure}

In this part, we show VEnhancer is able to achieve arbitrary up-sampling in time axis. Given two low-resolution key frames, we aim to up-sample them to high-resolution ones, and also interpolate several frames (ranging from $2$ to $4$) between them. As shown in Figure \ref{fig:multi_interp}, the results are consistent across frames, showing not flicking or distortions. Besides, the spatial quality has also been significantly improved. As shown in the last row, $5\times$ frame interpolation yields smooth frames with generated contents: the shadow in the right leg is changing, showing a very natural transition. This indicates that diffusion-based frame interpolation has great capability in both motion and content generation.

\end{document}